\pdfoutput=1

\documentclass[11pt]{article}

\usepackage[preprint]{acl}

\usepackage{times}
\usepackage{latexsym}

\usepackage[T1]{fontenc}

\usepackage[utf8]{inputenc}

\usepackage{microtype}

\usepackage{inconsolata}

\usepackage{graphicx}

\usepackage{booktabs}
\usepackage{multirow}
\usepackage{amssymb}
\usepackage{placeins}

\title{Active Learning for Finding a Needle in a Haystack}

\author{
 \textbf{Emily Silcock\textsuperscript{1},
 Simon Löwe\textsuperscript{2}}
\\
 \textsuperscript{1}Harvard University,
 \textsuperscript{2}Burning Glass Institute
\\
 \small{
   \textbf{Correspondence:} \href{mailto:emilysilcock@fas.harvard.edu}{emilysilcock@fas.harvard.edu}
 }
}

\begin{document}
\maketitle
\begin{abstract}
Training a classifier to find a rare class of documents, such as hate speech directed at a particular religious group, requires labelled examples that are expensive to obtain when the class makes up only a small fraction of a corpus. Active learning reduces this cost, but it must start from an initial labelled set that contains examples of the target class. Random sampling cannot supply these at reasonable cost, so they must be selected in some other way, such as keyword search, and any such selection biases the initial set towards part of the target class. We evaluate whether active learning can correct this bias. On AG News and Measuring Hate Speech, at a prevalence of around 1\%, we compare random sampling with eight active learning strategies, starting from initial sets that cover or omit known subgroups of the target class, and measure recall on the omitted subgroups as well as F1. On AG News, almost every strategy recovers the omitted subgroup. On Measuring Hate Speech, uncertainty-based strategies recover less of the omitted subgroups than random sampling, whereas discriminative active learning recovers 55\% of them without a significant loss of F1. Aggregate metrics conceal these differences.
\end{abstract}

\section{Introduction}
\label{sec:intro}

Many text classification tasks involve identifying a rare class of documents within a large corpus. Examples include finding news articles about a specific policy, detecting hate speech directed at a particular religious group, or identifying customer reviews that report a specific product fault. In each case, the target class may make up only a small fraction of the corpus, often 1\% or less. We refer to these as \emph{needle-in-a-haystack} problems.

Training a classifier for such a task requires labelled examples of the target class, which are costly to obtain when the class is rare. At 1\% prevalence, for example, labelling 1{,}000 randomly sampled documents yields around ten examples of the target class. Random sampling is therefore a prohibitively expensive way to obtain training data.

Active learning aims to reduce labelling costs by choosing which documents to label. A model trained on the labels collected so far selects the documents to be labelled next, typically those it expects to be most informative. The model is then retrained on the expanded labelled set, and the process repeats until the labelling budget is exhausted. Active learning should be most valuable for needle-in-a-haystack problems, where random sampling is least efficient.

However, active learning requires an \emph{initial set} of labelled documents that contains examples of the target class. Since random sampling cannot provide these at reasonable cost, they are typically found using keyword searches or similar heuristics. Initial sets obtained in this way are biased towards the part of the target class that the heuristic captures, and the direction of this bias is difficult to anticipate. For example, when searching for news coverage of contraception, ``the pill'' is too ambiguous to be a useful keyword, whereas ``birth control'' appears precise. In practice, however, coverage using the term ``birth control'' is heavily skewed towards Catholic perspectives on contraception. Random sampling is the only way to obtain an unbiased initial set, and it is precisely what the rarity of the target class rules out.

We therefore require two things of active learning in this setting. First, it should produce a high-quality classifier. Second, it should counteract the bias of the initial set by identifying the parts of the target class that it did not cover. These goals are distinct. Aggregate metrics such as F1 are dominated by the largest parts of the target class, so a classifier can perform well on them while systematically missing smaller parts of the class. This matters when the classifier is used to measure quantities of interest, as is common in the social sciences. If its errors are concentrated in particular parts of the target class, they bias downstream estimates \citep{fong2021machine, egami2023dsl}, and the statistical methods that correct for such errors require labelled documents sampled at random, which the rarity of the target class rules out \citep{carlson2025unifying}.

In this paper, we evaluate how well existing active learning methods meet these two goals on needle-in-a-haystack problems with biased initial sets. We construct binary classification tasks with a target-class prevalence of approximately 1\% from AG News \citep{zhang2015character} and the UCB Measuring Hate Speech corpus \citep{sachdeva2022measuring}, which we refer to as Measuring Hate Speech. Each task is run from both an unbiased initial set and a biased initial set drawn from a single subgroup of the target class. In Measuring Hate Speech, for example, the biased initial set contains only hate speech targeting Muslims, and hate speech targeting other religions must be found through active learning. Since subgroup labels are available, we measure directly how much of each subgroup unrepresented in the initial set is recovered. We compare random sampling with eight active learning strategies drawn from the main families of methods: uncertainty-based, diversity-based and representativeness-based methods, and methods that combine them. Each configuration is repeated over 25 runs.

We find that a biased initial set matters more for some tasks than others. On AG News, most strategies recover the unrepresented subgroup, and uncertainty-based strategies achieve among the highest F1. On Measuring Hate Speech, where the target class is harder to identify, the choice of strategy matters much more. With a biased initial set, the classifier trained with least-confidence sampling identifies 2\% of test documents in the unrepresented subgroups, compared with 55\% for discriminative active learning \citep[DAL;][]{DBLP:journals/corr/abs-1907-06347}, whose F1 is not significantly different from that of the best strategy. DAL also achieves the highest F1 with an unbiased initial set. The class-balanced variants of DAL improve F1 on AG News, but recover less of the unrepresented subgroups than DAL on Measuring Hate Speech. Precision on Measuring Hate Speech remains below 0.2 for every strategy, so the problem is far from solved.

\section{Related Work}
\label{sec:related}

Active learning strategies fall into two main families: uncertainty-based strategies, and diversity- or representativeness-based strategies \citep{settles2009active, zhang-etal-2022-survey}. Comparisons with fine-tuned transformers find that no strategy is consistently best \citep{ein-dor-etal-2020-active, schroder-etal-2022-revisiting, margatina-etal-2022-importance}. However, these comparisons use target classes with a prevalence of at least 5--10\% \citep{ein-dor-etal-2020-active, kirk-etal-2022-data}. This is an order of magnitude higher than the prevalence we study.

At a prevalence of 1\% or below, active learning has mainly been evaluated with earlier models \citep{DBLP:journals/corr/LewisG94, attenberg2010label}. It has also been studied in technology-assisted review, where the aim is to rank documents until most relevant ones are found, rather than to train a classifier \citep{cormack2014evaluation, yang2022goldilocks}. Two studies use transformers at this prevalence. \citet{mussmann-etal-2020-importance} show that uncertainty sampling far outperforms fixed data collection for pairwise tasks with extremely rare positives. \citet{lesci-vlachos-2024-anchoral} reduce the cost of selecting from very large pools by scoring only documents similar to those already labelled. Like most evaluations of active learning, these studies report aggregate metrics. These show how well a classifier performs overall, but not which parts of the target class it finds.

Knowing which parts of the target class are found is important when the initial set is biased. With earlier models, active learning can fail to query clusters that are missing from the initial labelled set, because the learner is confident about their labels \citep{dasgupta2011two, tomanek-etal-2009-proper}. \citet{attenberg2011inactive} observe this within a rare target class, and warn that aggregate metrics conceal it. It is unclear whether this carries over to fine-tuned transformers. Pretrained representations may already separate the missing clusters, or the fine-tuned classifier may assign them confidently to the wrong class. \citet{ein-dor-etal-2020-active} come closest to testing this. They draw the initial set from a keyword query, and find that active learning with BERT then improves recall most. However, they measure recall on the target class as a whole, so they cannot tell whether active learning found the parts of the target class that the query missed. We construct initial sets that omit known subgroups of the target class, and measure recall on those subgroups directly.

As Section~\ref{sec:intro} notes, a classifier whose errors are concentrated in parts of the target class biases the measurements made with it. Prediction-powered inference \citep{angelopoulos2023ppi} and design-based supervised learning \citep{egami2023dsl} correct this bias using a labelled sample, provided that every document has a known, positive probability of being labelled \citep{carlson2025unifying}. Selecting documents by keyword search violates this condition \citep{carlson2025unifying, dell2026measurement}, and for a rare target class a sample that meets it is expensive to obtain. Reducing the bias of the classifier itself is therefore complementary to these corrections.

\section{Problem Setup}
\label{sec:problem}

\subsection{Needle-in-a-haystack classification}
\label{sec:needle}

We consider a corpus $\mathcal{D}$ of documents, each with an unobserved binary label $y \in \{0, 1\}$, where $y = 1$ indicates membership of the target class. The prevalence of the target class, $\pi = \Pr(y = 1)$, is small. We focus on $\pi \approx 1\%$, and in many applications it is lower still. The aim is to train a classifier that identifies target-class documents in $\mathcal{D}$ and in new documents drawn from the same distribution.

Labels are obtained from an annotator, up to a budget of $B$ documents. In pool-based active learning, labelling proceeds in rounds, starting from an initial set $\mathcal{L}_0$ of labelled documents. In round $t$, a classifier $f_t$ is trained on the labelled set $\mathcal{L}_t$. A query strategy then uses $f_t$ to select a batch $\mathcal{Q}_{t+1}$ of $b$ documents from the unlabelled pool $\mathcal{D} \setminus \mathcal{L}_t$. These documents are labelled and added to the labelled set, $\mathcal{L}_{t+1} = \mathcal{L}_t \cup \mathcal{Q}_{t+1}$. After $T = B / b$ rounds, the final classifier $f_T$ is trained on $\mathcal{L}_T$. Random sampling is the special case in which the query strategy selects documents uniformly at random from the pool.

\subsection{Biased initial sets}
\label{sec:biased}

In practice, the initial set must contain target-class documents. Without them, the first classifier cannot learn what the target class looks like, and query strategies, which rely on this classifier, have nothing to work from. At low prevalence, random sampling cannot supply these documents at reasonable cost. Instead, they are found by selecting documents on their observable features, typically with a keyword search. Any such method draws disproportionately from the parts of the target class that share those features, so the initial set is biased. No choice of heuristic avoids this, since random sampling is the only unbiased way to find target-class documents. The direction of the bias is also difficult to anticipate, as the contraception example in Section~\ref{sec:intro} illustrates.

This bias matters because active learning can perpetuate it. The first classifier learns to recognise the kinds of target-class documents in the initial set, and may assign low scores to other kinds. A query strategy that relies on these scores may never select the other kinds, in which case later classifiers inherit the same bias. Avoiding this requires a strategy to select documents that the current classifier does not regard as likely members of the target class.

The bias cannot be corrected directly, because the practitioner does not know which parts of the target class the initial set misses. If they did, they could search for those parts. Whether the bias is corrected therefore depends on the query strategy.

As a stylised representation of this bias, suppose that the target class is divided into a set of subgroups $\mathcal{G}$, and write $g(x) \in \mathcal{G}$ for the subgroup to which target-class document $x$ belongs. For hate speech directed at religious groups, for example, the subgroups might be the religions targeted. For news coverage of contraception, they might be the perspectives from which it is discussed. A biased initial set contains target-class documents from only some subgroups, which we call the \emph{represented} subgroups, $\mathcal{G}_R \subset \mathcal{G}$. We call the remainder the \emph{unrepresented} subgroups, $\mathcal{G}_U = \mathcal{G} \setminus \mathcal{G}_R$. Correcting the bias then means finding target-class documents in $\mathcal{G}_U$. Real target classes are rarely divided into discrete subgroups, but this representation makes the bias, and its correction, measurable.

\section{Methods}
\label{sec:methods}

We compare random sampling, which selects documents uniformly at random from the pool, with eight active learning strategies. They come from the two main families of active learning methods: uncertainty-based strategies, which select documents about which the current classifier is unsure, and diversity- and representativeness-based strategies, which select documents that differ from those already labelled or that are typical of the pool. Two of the eight are class-balanced variants of one of these strategies, discriminative active learning. Every strategy selects a batch of 100 documents in each round. Appendix~\ref{sec:appendix-strategies} gives implementation details.

\subsection{Uncertainty-based strategies}
\label{sec:uncertainty}

Least confidence \citep[LC;][]{DBLP:journals/corr/LewisG94} selects the documents whose most probable class has the lowest predicted probability. With two classes, these are the documents whose predicted probability of the target class is closest to 0.5.

BALD \citep{houlsby2011bayesian, pmlr-v70-gal17a} selects the documents on which the classifier's predictions disagree most across stochastic forward passes with dropout. This disagreement estimates how much the document's label would reveal about the classifier's parameters.

Both strategies select documents close to the current decision boundary. They therefore rely on the classifier's predicted probabilities, which a biased initial set may distort (Section~\ref{sec:biased}).

\subsection{Diversity- and representativeness-based strategies}
\label{sec:diversity}

Core-Set \citep{sener2018active, DBLP:journals/corr/abs-1711-00941} selects documents that cover the pool. It repeatedly selects the unlabelled document farthest from its nearest labelled or already-selected document in the classifier's embedding space, so that every document in the pool is close to one that has been labelled.

BADGE \citep{ash2020deep} is designed to combine uncertainty and diversity. It represents each document by the gradient of the training loss with respect to the classifier's final layer, whose magnitude reflects uncertainty, and selects a diverse set of these gradient embeddings using k-means++ initialisation.

Contrastive active learning \citep[CAL;][]{margatina-etal-2021-active} selects documents that resemble labelled documents but receive different predictions. For each unlabelled document, it computes the mean Kullback--Leibler divergence between the document's predicted class distribution and those of its ten nearest labelled neighbours in embedding space, and selects the documents with the largest divergence.

Discriminative active learning \citep[DAL;][]{DBLP:journals/corr/abs-1907-06347} treats selection as a classification problem. It trains a discriminator to distinguish labelled from unlabelled documents, using the classifier's representation of each document, and selects the unlabelled documents that the discriminator most confidently identifies as unlabelled, that is, those least like the labelled set. To avoid selecting many similar documents at once, the batch is selected in 20 sub-batches, and the discriminator is retrained after each with the documents selected so far treated as labelled. We implement DAL as specified by its authors. The discriminator is a multilayer perceptron with three hidden layers of 256 units, trained on the input to the classifier's final layer until it classifies more than 98\% of its training documents correctly.

Core-Set and DAL select documents using only the classifier's representation of each document, not its predicted probabilities. BADGE and CAL use both.

\subsection{Class-balanced DAL}
\label{sec:cbdal}

At low prevalence, almost every unlabelled document belongs to the non-target class. A strategy that selects documents unlike the labelled set may therefore select mostly non-target documents, which tell the classifier little about the target class. A natural extension of DAL is to balance its pool between documents that the classifier predicts to belong to the target class and documents that it predicts do not. We test two such variants. Both use the same discriminator as DAL and differ from it only in the pool from which they select.

The first, class-balanced DAL v1, splits the unlabelled pool into documents predicted to belong to the target class and documents predicted not to, and runs DAL separately on each to select half of the batch from each. In each half, the discriminator distinguishes the unlabelled documents from the labelled documents of the corresponding class.

Class-balanced DAL v2 balances the pool by subsampling instead. It randomly subsamples the larger of the two predicted pools to the size of the smaller, and runs DAL once on the combined pool against the full labelled set, so the batch need not be divided evenly between the two.

Both variants use the classifier's predictions to divide the pool. Unlike DAL, they therefore depend on the classifier's estimate of where the target class lies.

\section{Experimental Setup}
\label{sec:exp-setup}

\subsection{Datasets}
\label{sec:datasets}

\begin{table*}[t]
\centering
\small
\begin{tabular}{@{}llllrrr@{}}
\toprule
Dataset & Initial set & Target class & \shortstack[l]{Initial-set target\\documents from} & Train size & Test size & \shortstack[r]{\% target\\(train)} \\
\midrule
\multirow{2}{*}{AG News}         & Unbiased & World                  & World             & 90{,}909 & 5{,}757 & 1.00\% \\
                                  & Biased   & World and Sports       & World             & 60{,}606 & 3{,}838 & 1.00\% \\
\cmidrule(lr){2-7}
\multirow{2}{*}{\shortstack[l]{Measuring\\Hate Speech}}     & Unbiased & Religion targeted      & All religions     & 31{,}587 & 7{,}901 & 1.34\% \\
                                  & Biased   & Religion targeted      & Muslims           & 31{,}587 & 7{,}901 & 1.34\% \\
\bottomrule
\end{tabular}
\caption{Tasks used in the experiments. ``Initial-set target documents from'' gives the part of the target class from which the 50 target-class documents in the initial set are drawn. The AG News target class is down-sampled separately for each random seed (Section~\ref{sec:training}); the sizes shown are the same for every seed.}
\label{tab:datasets}
\end{table*}

We use two English text classification corpora, AG News \citep{zhang2015character} and the UCB Measuring Hate Speech corpus \citep{sachdeva2022measuring}, which we refer to as Measuring Hate Speech.

AG News consists of news articles labelled with one of four topics: World, Sports, Business and Sci/Tech. It contains 120{,}000 training and 7{,}600 test articles, divided evenly between the topics. It has no annotations below the level of topic.

Measuring Hate Speech consists of 39{,}565 social media comments, each rated by several annotators. The annotations include a continuous hate speech score and indicators of the groups that each comment targets, including eight religious groups. The target class is hate speech directed at a religious group. We average the annotations of each comment across annotators, and assign the comment to the target class if its average hate speech score exceeds 1 and all annotators agree that it targets a religion. The subgroups are the religions targeted: atheists, Buddhists, Christians, Hindus, Jews, Mormons, Muslims and other religions. Muslims are the largest subgroup. The corpus is released without a test split, so we form a stratified 80/20 split. In the test set, subgroup sizes range from one comment (Mormons) to 47 (Muslims). Appendix~\ref{sec:appendix-datasets} gives further details of the preprocessing.

\subsection{Target classes and initial sets}
\label{sec:datapprep}

We run every experiment from two kinds of initial set. With an unbiased initial set, the target-class documents are a random sample of the whole target class. This is not obtainable in practice, but it serves as a control. With a biased initial set, all target-class documents in the initial set come from a single subgroup, the represented subgroup. This is the most extreme form of the bias described in Section~\ref{sec:biased}, and it makes the distinction between represented and unrepresented subgroups unambiguous. Comparing the two settings separates the effect of the bias from the effect of low prevalence alone. Subgroup labels are used only to construct the initial sets and to evaluate the classifiers, and are never available to the query strategies.

In both settings, the initial set contains 100 documents: 50 from the target class and 50 drawn at random from the non-target class. The settings differ only in how the 50 target-class documents are drawn. Including equal numbers of each class gives the first classifier a usable number of target-class examples, as a practitioner who had searched for them would have.

In Measuring Hate Speech, the target class is the same in both settings. It makes up 1.34\% of the training set, close to the prevalence we study, so we use it at its natural rate. With an unbiased initial set, the target-class documents in the initial set are drawn from all eight subgroups. With a biased initial set, Muslims are the represented subgroup. We choose the largest subgroup because it is the kind of target-class document that a keyword search is most likely to find. The other seven subgroups are unrepresented. They contain 240 of the 424 target-class comments in the training set and 63 of the 110 in the test set.

AG News has no subgroups within a topic, so we construct the target class differently in each setting. With an unbiased initial set, the target class is World, and the non-target class consists of the other three topics. With a biased initial set, the target class is World and Sports combined, with World as the represented subgroup and Sports as the unrepresented subgroup, and the non-target class consists of Business and Sci/Tech. In both settings, we down-sample the target class at random until it makes up 1\% of the training and test sets, keeping all non-target articles, so World and Sports each make up about half of the biased target class. Because the target classes differ, the two AG News settings are a less direct comparison than the two Measuring Hate Speech settings, in which only the initial set changes.

The AG News test set contains 57 target-class articles in the unbiased setting and 38 in the biased setting. The down-sampling depends on the random seed (Section~\ref{sec:training}), so the articles kept, including the number of Sports articles among the 38 (between 13 and 24), vary across repetitions of the experiment but are the same for every method. Table~\ref{tab:datasets} summarises the resulting tasks, and Appendix~\ref{sec:appendix-datasets} gives further details.

\subsection{Model, budget and training}
\label{sec:training}

All methods use the same classifier, so they differ only in the documents they select. The classifier is DistilRoBERTa, a distilled version of RoBERTa \citep{sanh2019distilbert, liu2019roberta}, fine-tuned with a binary classification head on inputs truncated to 100 tokens. In each round, it is retrained from the pretrained weights on the full labelled set. We weight the training loss by inverse class frequency, since the class balance of the labelled set depends on the documents a method selects. We implement the methods with the small-text library \citep{schroder-etal-2023-small}, except for DAL and its variants, which we implement ourselves (Appendix~\ref{sec:appendix-strategies}). Appendix~\ref{sec:appendix-training} lists the hyperparameters.

Each experiment starts from the initial set of 100 documents and runs for ten rounds of 100 queries, so the final classifier is trained on 1{,}100 labelled documents. At this budget, random sampling would be expected to find around ten target-class documents beyond those in the initial set.

We repeat each combination of dataset, setting and method 25 times, with a different random seed each time, and refer to each repetition as a run. The seed determines the initial set, the AG News down-sampling, and the classifier's initialisation and training order. Within a run, all methods share the same seed, and therefore the same data and initial set, so comparisons between methods are paired by run. Classifier training on a GPU is not fully deterministic, however, so classifiers trained on identical labelled documents can differ slightly.

The number of training epochs is the only hyperparameter we tuned. We chose it from $\{3, 5, 8, 10\}$ using random sampling, with three runs per setting, and selected ten epochs, which gave the highest mean subgroup recall and an F1 close to the best (Appendix~\ref{sec:appendix-training}). The corpora have no separate validation sets, so the selection used the test sets. We tuned only random sampling, so the choice does not favour any active learning method. The tuning runs used the same random seeds as three of the 25 reported runs. All other hyperparameters, including those of the query strategies, are set to the values recommended by the methods' authors or to the small-text defaults.

\subsection{Evaluation}
\label{sec:metrics}

The two goals set out in Section~\ref{sec:intro} correspond to two metrics. Both are computed for the final classifier $f_T$ on the test set, in which each document $x$ has a label $y$ and, if it belongs to the target class, a subgroup $g(x)$ as defined in Section~\ref{sec:biased}.

The first goal is a high-quality classifier, which we measure with F1 on the target class. Accuracy is uninformative at low prevalence, since a classifier that assigns every document to the non-target class achieves an accuracy of $1 - \pi$.

The second goal is to correct the bias of the initial set. F1 does not capture this goal. It weights subgroups by their size, not by whether they are represented in the initial set, so a classifier can achieve a high F1 while missing unrepresented subgroups that make up a small share of the target class. Instead, we measure it with recall on unrepresented subgroups, the proportion of target-class documents in unrepresented subgroups that the classifier assigns to the target class:
\[
R_U = \Pr\left(f_T(x) = 1 \mid y = 1,\ g(x) \in \mathcal{G}_U\right).
\]

A high $R_U$ alone is not evidence that the bias has been corrected. A classifier that assigns more documents to the target class increases its recall on every subgroup, including the unrepresented ones, at the cost of precision. We therefore compare $R_U$ with recall on the represented subgroups, $R_R$, defined analogously. If a method corrects the bias, $R_U$ approaches $R_R$, whereas if it only assigns more documents to the target class, both increase together. We also report $R_U$ alongside F1, so that gains in $R_U$ that come at a large cost in precision are visible.

With an unbiased initial set, every subgroup is represented, so $R_U$ is not defined and we compare methods on F1 alone.

To compare methods, we take the method with the highest mean on each metric in each setting and test every other method against it, using a paired $t$-test across the 25 runs. We correct for the eight comparisons in each test with the Holm procedure \citep{holm1979simple}.

\section{Results}
\label{sec:results}

Table~\ref{tab:main} reports performance after the final round. We first describe the results with an unbiased initial set, then with a biased initial set, and then examine which documents each strategy selects.

\begin{table*}[t]
\centering
\small
\setlength{\tabcolsep}{5pt}
\begin{tabular}{@{}lc@{\hspace{10pt}}ccc@{\hspace{18pt}}c@{\hspace{10pt}}ccc@{}}
\toprule
 & \multicolumn{4}{c}{AG News} & \multicolumn{4}{c}{Measuring Hate Speech} \\
\cmidrule(lr){2-5}\cmidrule(l){6-9}
 & Unbiased & \multicolumn{3}{c}{Biased} & Unbiased & \multicolumn{3}{c}{Biased} \\
\cmidrule(lr){2-2}\cmidrule(lr){3-5}\cmidrule(lr){6-6}\cmidrule(l){7-9}
Method & F1 & F1 & $R_U$ & $R_R$ & F1 & F1 & $R_U$ & $R_R$ \\
\midrule
Random & 0.52 & 0.41 & 0.56 & 0.78 & 0.13 & \underline{0.13} & 0.17 & 0.49 \\
\midrule
LC & \underline{0.69} & \textbf{0.73} & \underline{0.90} & 0.68 & 0.05 & 0.06 & 0.02 & 0.10 \\
BALD & \underline{0.68} & \underline{0.68} & \underline{0.87} & 0.65 & 0.08 & 0.10 & 0.09 & 0.23 \\
\midrule
Core-Set & \underline{0.63} & 0.69 & \underline{0.90} & 0.63 & 0.11 & 0.10 & 0.16 & 0.30 \\
BADGE & \underline{0.66} & \underline{0.71} & \textbf{0.91} & 0.65 & 0.09 & \underline{0.12} & 0.10 & 0.22 \\
CAL & \underline{0.68} & 0.70 & \underline{0.86} & 0.62 & 0.07 & 0.12 & 0.14 & 0.20 \\
DAL & 0.58 & 0.63 & \underline{0.91} & 0.76 & \textbf{0.19} & \underline{0.16} & \textbf{0.55} & 0.53 \\
\midrule
CB-DAL v1 & 0.63 & 0.67 & \underline{0.91} & 0.68 & 0.14 & \textbf{0.18} & 0.37 & 0.34 \\
CB-DAL v2 & \textbf{0.70} & \underline{0.72} & \underline{0.90} & 0.63 & 0.11 & \underline{0.14} & 0.12 & 0.29 \\
\bottomrule
\end{tabular}
\caption{Test performance after the final round (1{,}100 labelled documents), mean over 25 runs. Unbiased and Biased refer to the initial set. $R_U$: recall on the subgroups unrepresented in the biased initial set; $R_R$: recall on the represented subgroup. Bold: best method in the column; underlined: not significantly worse than the best (paired $t$-test across runs, Holm-corrected over the eight comparisons in the column, $p \geq 0.05$). $R_R$ is not tested. Standard errors are in Table~\ref{tab:main-se}.}
\label{tab:main}
\end{table*}

\subsection{Unbiased initial set}
\label{sec:results-unbiased}

On AG News, most strategies improve substantially on random sampling. Random sampling reaches an F1 of 0.52, whereas class-balanced DAL v2, LC, CAL, BALD, BADGE and Core-Set reach 0.63--0.70 and are not significantly different from one another. DAL reaches 0.58. Its recall is the highest of any strategy, but its precision is the lowest apart from random sampling (Appendix~\ref{sec:appendix-results}).

On Measuring Hate Speech, every strategy performs poorly, and the ranking differs. DAL achieves the highest F1, 0.19, and is significantly better than every other strategy. Its advantage comes from recall: it identifies 55\% of the target class in the test set, compared with at most 35\% for any other strategy. The uncertainty-based strategies perform worst. LC identifies 5\% of the target class and has an F1 of 0.05, below random sampling. Precision is below 0.17 for every strategy.

\subsection{Biased initial set}
\label{sec:results-biased}

\begin{figure*}[t]
\centering
\includegraphics[width=\textwidth]{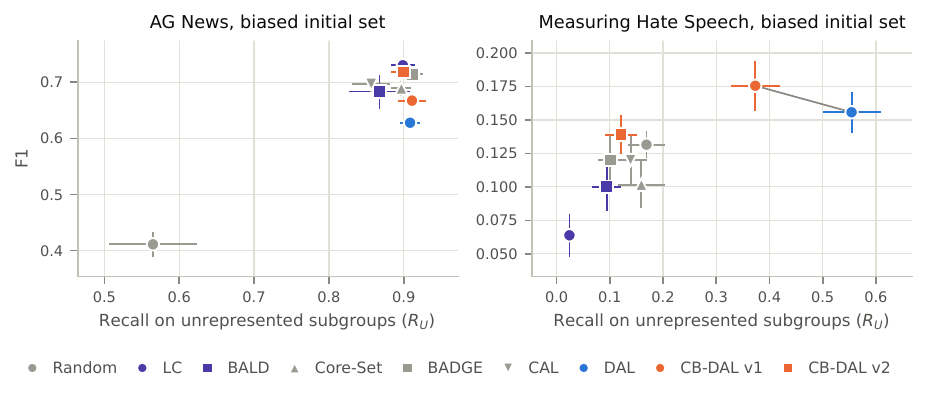}
\caption{Recall on the subgroups unrepresented in the initial set ($R_U$) against F1 after the final round, with a biased initial set. Points are means over 25 runs, with standard errors. The grey line joins methods not outperformed on both metrics by any other method.}
\label{fig:tradeoff}
\end{figure*}

\begin{figure*}[t]
\centering
\includegraphics[width=\textwidth]{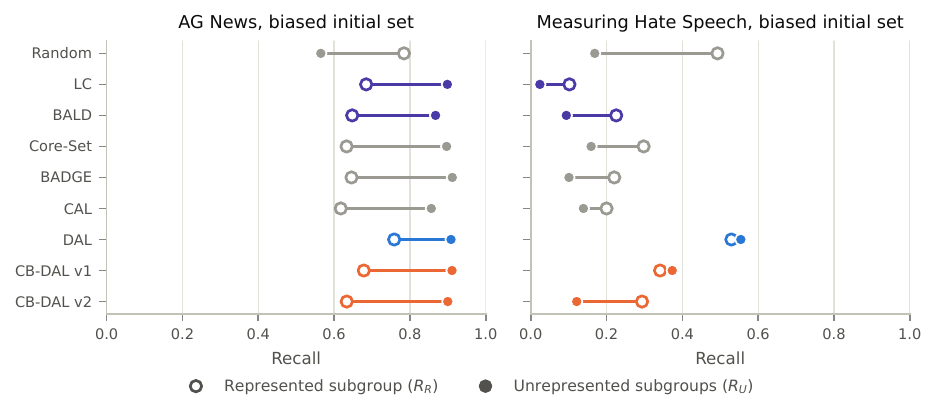}
\caption{Recall on the represented subgroup ($R_R$, open) and on the unrepresented subgroups ($R_U$, filled) after the final round, with a biased initial set, mean over 25 runs. A method that corrects the bias of the initial set closes the gap between the two.}
\label{fig:gap}
\end{figure*}

On AG News, the biased initial set has little effect. Every strategy except random sampling identifies 86--91\% of the unrepresented subgroup, Sports, and these strategies are not significantly different from one another (Figure~\ref{fig:tradeoff}). For every strategy except random sampling, recall on the unrepresented subgroup is higher than on the represented subgroup, World (Figure~\ref{fig:gap}). Sports articles appear easy to identify once a few have been labelled. F1 is similar to that with an unbiased initial set, with LC, class-balanced DAL v2, BADGE and BALD the best.

\begin{figure*}[t]
\centering
\includegraphics[width=\textwidth]{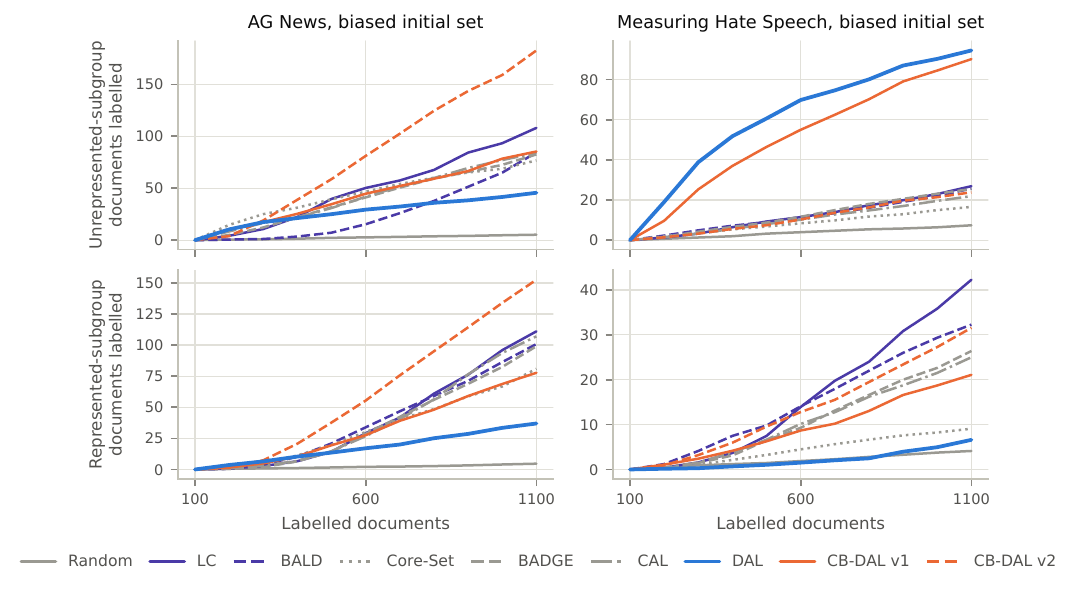}
\caption{Target-class documents from the unrepresented subgroups (top) and from the represented subgroup (bottom) selected for labelling, cumulatively over rounds, with a biased initial set. Mean over 25 runs; excludes the 50 target-class documents in the initial set.}
\label{fig:recovery}
\end{figure*}

On Measuring Hate Speech, the bias persists for most strategies. Recall on the unrepresented subgroups ranges from 0.02 for LC to 0.55 for DAL, which is significantly higher than for every other strategy. DAL and class-balanced DAL v1 are the only strategies whose recall on the unrepresented subgroups is at least as high as on the represented subgroup (Figure~\ref{fig:gap}). For every other strategy, the classifier identifies a smaller share of the unrepresented subgroups than of the represented one. Every strategy other than DAL and class-balanced DAL v1 recovers less of the unrepresented subgroups than random sampling does.

With a biased initial set on Measuring Hate Speech, the two goals are compatible. Class-balanced DAL v1 achieves the highest F1, 0.18, but DAL, class-balanced DAL v2, BADGE and random sampling are not significantly worse. DAL therefore achieves by far the highest recall on the unrepresented subgroups at no significant cost in F1. On AG News, by contrast, DAL recovers the unrepresented subgroup as well as the best strategies but has a lower F1.

\subsection{Which documents are selected}
\label{sec:results-selection}

Figure~\ref{fig:recovery} shows how many documents from each subgroup of the target class each strategy selects for labelling. On Measuring Hate Speech, DAL selects 101 target-class documents over ten rounds, of which 94 are from the unrepresented subgroups. These subgroups make up 64\% of the target-class documents in the pool, but 93\% of those selected by DAL. Class-balanced DAL v1 behaves similarly. LC selects 69 target-class documents, but only 39\% of them are from the unrepresented subgroups. It concentrates on documents similar to the represented subgroup, as Section~\ref{sec:biased} anticipates for a strategy that relies on the classifier's predicted probabilities.

On AG News, every strategy selects documents from the two subgroups in roughly equal numbers, in proportion to their share of the pool. The strategies differ instead in how many target-class documents they select. DAL selects 83, compared with 219 for LC and 335 for class-balanced DAL v2, which is consistent with its lower precision on this dataset. Balancing the pool between predicted target-class and predicted non-target documents increases the number of target-class documents selected on AG News, but not on Measuring Hate Speech, where class-balanced DAL v2 selects 55. On Measuring Hate Speech, only around 10\% of the unlabelled documents that the classifier predicts to be in the target class actually are, compared with 50--60\% on AG News, so balancing on these predictions adds few target-class documents.

\section{Discussion}
\label{sec:discussion}

Our results show that a biased initial set does not always matter, but that when it does, the choice of strategy determines whether the bias is corrected. On AG News, almost every strategy recovers the unrepresented subgroup. On Measuring Hate Speech, recall on the unrepresented subgroups varies by a factor of more than twenty across strategies. With two datasets we cannot identify what distinguishes them, but the precision of the classifiers suggests one explanation. On AG News, the classifier separates the target class well, so the first articles labelled from the unrepresented subgroup are enough for it to learn the subgroup. On Measuring Hate Speech, average precision is 0.12 or less in every round, so the classifier's estimate of where the target class lies remains poor. This is the setting in which the strategy matters most, and it is also the setting that a practitioner facing a rare and loosely defined target class is likely to encounter.

The uncertainty-based strategies fail in the way that earlier work describes for simpler models \citep{dasgupta2011two, attenberg2011inactive}. A classifier trained on the represented subgroup assigns documents from the other subgroups confidently to the non-target class, so they are rarely selected, and the labelled set grows mainly around the represented subgroup. Section~\ref{sec:related} noted that pretrained representations might prevent this. Our results suggest that they do not: with a fine-tuned transformer, least-confidence sampling recovers less of the unrepresented subgroups than random sampling. BADGE, which selects diverse documents but weights them by the classifier's uncertainty, fails in the same way.

DAL avoids this failure. It selects documents that are unlike the labelled set without using the classifier's predicted probabilities, so its selections do not depend on the classifier's biased estimate of where the target class lies, and it selects documents from the unrepresented subgroups disproportionately. Not using predicted probabilities is not sufficient, however. Core-Set also selects documents far from the labelled set without using them, but on Measuring Hate Speech it selects few target-class documents and recovers no more of the unrepresented subgroups than random sampling. Our experiments do not explain this difference. DAL's approach also has a cost. Most documents in the pool belong to the non-target class, and on AG News DAL selects far fewer target-class documents than the other strategies, which lowers its precision on a task where the classifier could otherwise separate the target class well. Balancing the pool on the classifier's predictions addresses this cost only when the predictions are informative. On Measuring Hate Speech they are not: class-balanced DAL v1 recovers less of the unrepresented subgroups than DAL, and v2 recovers less than random sampling.

These results qualify the finding of \citet{ein-dor-etal-2020-active} that active learning from a keyword-based initial set improves recall. At a prevalence of around 1\%, higher recall can come either from recovering the parts of the target class that the initial set missed or from finding more of the parts it covered, and which of these occurs depends on the strategy. Aggregate metrics cannot distinguish the two, which is why we measure recall on the unrepresented subgroups directly.

In practice, subgroup labels are unavailable, so a practitioner cannot measure how much of the bias has been corrected. Our results suggest two precautions. First, when the target class is rare and difficult to identify, uncertainty-based strategies risk reinforcing the bias of the initial set, and DAL was the most reliable strategy in our experiments. Second, a classifier that performs well on aggregate metrics may still miss parts of the target class. Any bias that remains passes into measurements made with the classifier, and, as discussed in Section~\ref{sec:related}, it cannot easily be corrected afterwards when the labelled documents were not selected at random.

\FloatBarrier

\section{Conclusion}
\label{sec:conclusion}

Many text classification tasks require finding a target class that makes up only a small fraction of a corpus. Active learning should be most valuable in this setting, but its initial labelled set cannot be found by random sampling, and any other way of finding it is biased. We compared eight active learning strategies at a prevalence of around 1\%, from biased and unbiased initial sets, measuring both F1 and recall on the parts of the target class that the initial set missed. Whether the bias matters depends on the task. When it does, uncertainty-based strategies reinforce it, whereas discriminative active learning corrects it without a significant loss of F1. No strategy is best on every task, and on the harder task every strategy achieves low precision. Evaluations of active learning for rare classes should therefore report which parts of the target class a classifier finds, not only aggregate metrics.

\section*{Limitations}

Our evidence comes from two datasets, and the bias of the initial set matters for only one of them. We cannot therefore say which properties of a task determine whether the bias persists, and the ranking of strategies may differ on other tasks.

The biased initial sets are stylised. In each, all target-class documents come from a single subgroup, and subgroups are discrete. Initial sets found by keyword search in practice are likely to be biased less completely, and towards parts of the target class that do not correspond to annotated subgroups.

Several subgroups are small. In the Measuring Hate Speech test set, subgroup sizes range from one to 47 comments, and the AG News test set contains between 13 and 24 Sports articles in each run. Recall on individual subgroups is therefore noisy, which is why we report recall on the unrepresented subgroups combined, averaged over 25 runs.

The two AG News settings use different target classes, so comparing them confounds the bias of the initial set with the change in target class. The two Measuring Hate Speech settings do not have this problem.

We use a single classifier, DistilRoBERTa, a single labelling budget of 1{,}100 documents, and binary classification in English. The strategies use the settings recommended by their authors or the defaults of the library we use, and the number of training epochs was chosen on the test sets using random sampling only. Different classifiers, budgets or settings may change the results.
\section*{Acknowledgments}

We thank Lukas Fesser for fruitful discussions.

\FloatBarrier

\bibliography{custom}

\appendix
\onecolumn

\section{Query strategy details}
\label{sec:appendix-strategies}

We use the implementations in small-text 1.4.1 \citep{schroder-etal-2023-small} for random sampling, LC, BALD, Core-Set, BADGE and CAL, with the settings in Table~\ref{tab:strategies}. The BADGE implementation in this version is incorrect, as noted in the library's errata, so we replace its computation of the gradient embeddings and its k-means++ initialisation with the corrected versions from small-text 2.0.0.dev4. Core-Set, BADGE and CAL represent each document by the mean of the classifier's final-layer token representations.

\begin{table}[htb]
\centering
\small
\begin{tabular}{@{}lp{0.62\columnwidth}@{}}
\toprule
Strategy & Settings \\
\midrule
BALD & 10 dropout samples \\
Core-Set & Greedy $k$-centre; Euclidean distance between unit-normalised embeddings \\
BADGE & k-means++ initialisation on gradient embeddings \\
CAL & 10 nearest labelled neighbours; unit-normalised embeddings \\
\bottomrule
\end{tabular}
\caption{Settings of the small-text query strategies. LC and random sampling have no settings.}
\label{tab:strategies}
\end{table}

We implement DAL in PyTorch, following the authors' reference implementation.\footnote{\texttt{github.com/dsgissin/\allowbreak DiscriminativeActiveLearning}} Each document is represented by the input to the classifier's final layer, which for our classifier is the transformed representation of the first token. These representations are computed once per round, with the classifier fixed, and are not rescaled. The discriminator has three hidden layers of 256 rectified linear units and is trained with Adam (learning rate $10^{-3}$, batch size 32) on a cross-entropy loss weighted by inverse class frequency. Training stops once accuracy on the discriminator's training documents exceeds 0.98, or after 1{,}000 epochs. In each of the 20 sub-batches of five documents, the unlabelled documents used to train the discriminator are a random sample of at most ten times the number of labelled documents, and a new discriminator is trained from scratch.

The two class-balanced variants use the same discriminator as DAL. In v1, if one predicted pool contains no more than half a batch, all of its documents are selected and the remainder of the batch is selected by DAL from the other pool. In v2, if either predicted pool is empty, DAL is run on the full unlabelled pool.

\section{Dataset details}
\label{sec:appendix-datasets}

\paragraph{AG News.} We use the \texttt{ag\_news} release on the Hugging Face Hub, with its original training and test splits of 120{,}000 and 7{,}600 articles. The non-target topics are kept in full: 90{,}000 training and 5{,}700 test articles in the unbiased setting, and 60{,}000 and 3{,}800 in the biased setting. The target class is then down-sampled to 1\% of each split, leaving 909 training and 57 test articles in the unbiased setting, and 606 training and 38 test articles in the biased setting.

\paragraph{Measuring Hate Speech.} We use the \texttt{ucberkeley-dlab/measuring-hate-speech} release on the Hugging Face Hub, which contains 135{,}556 ratings of 39{,}565 comments. For each comment, we average across annotators the hate speech score, the indicator for whether the comment targets religion, and the indicators for each of the eight religious groups. The target class is then defined as in Section~\ref{sec:datasets}.

We assign target-class comments to subgroups using the same criterion, so that a comment belongs to a religion's subgroup only if all annotators marked it as targeting that religion. A comment that targets more than one religion is assigned to the least frequently targeted of them, which keeps the smallest subgroups as large as possible. We drop the 77 target-class comments that meet this criterion for no individual religion, leaving 39{,}488 comments.

The release has no test split, so we divide the comments 80/20 into training and test sets. The split is stratified by subgroup, with comments outside the target class forming one further stratum, and is the same in every run. The training set contains 31{,}587 comments, of which 424 are in the target class, and the test set contains 7{,}901, of which 110 are in the target class. The test-set subgroups contain 47 comments targeting Muslims, 26 targeting Jews, 19 targeting Christians, 7 targeting other religions, 4 targeting Buddhists, 3 targeting atheists, 3 targeting Hindus and 1 targeting Mormons.

\section{Training details}
\label{sec:appendix-training}

Table~\ref{tab:hyperparameters} lists the hyperparameters of the classifier, which are shared by every method. Table~\ref{tab:epochs} reports the epoch tuning described in Section~\ref{sec:training}. Ten epochs gave the highest mean subgroup recall across the four settings (0.514, against 0.506 for eight epochs and 0.485 for five) and a mean F1 within 0.01 of the best (0.334, against 0.340 for five epochs).

\begin{table}[htb]
\centering
\small
\begin{tabular}{@{}ll@{}}
\toprule
Hyperparameter & Value \\
\midrule
Pretrained model & \texttt{distilroberta-base} \\
Maximum input length & 100 tokens \\
Training epochs per round & 10 \\
Batch size & 32 \\
Learning rate & $5 \times 10^{-5}$ \\
Loss weighting & Inverse class frequency \\
Initialisation each round & Pretrained weights \\
\bottomrule
\end{tabular}
\caption{Classifier hyperparameters.}
\label{tab:hyperparameters}
\end{table}

\begin{table}[htb]
\centering
\small
\setlength{\tabcolsep}{3pt}
\begin{tabular}{@{}llrrrr@{}}
\toprule
& & \multicolumn{4}{c}{Epochs} \\
\cmidrule(lr){3-6}
Dataset & Initial set & 3 & 5 & 8 & 10 \\
\midrule
\multicolumn{6}{@{}l}{\emph{F1}} \\
AG News & Unbiased & 0.51 & 0.57 & 0.47 & 0.55 \\
AG News & Biased & 0.46 & 0.51 & 0.44 & 0.47 \\
Measuring Hate Speech & Unbiased & 0.15 & 0.18 & 0.10 & 0.17 \\
Measuring Hate Speech & Biased & 0.11 & 0.10 & 0.19 & 0.15 \\
\midrule
\multicolumn{6}{@{}l}{\emph{Mean subgroup recall}} \\
AG News & Unbiased & 0.75 & 0.74 & 0.84 & 0.77 \\
AG News & Biased & 0.62 & 0.71 & 0.68 & 0.68 \\
Measuring Hate Speech & Unbiased & 0.26 & 0.26 & 0.25 & 0.41 \\
Measuring Hate Speech & Biased & 0.16 & 0.23 & 0.25 & 0.20 \\
\bottomrule
\end{tabular}
\caption{Epoch tuning: final-round test performance of random sampling, mean over three runs (two runs for AG News unbiased at three epochs and Measuring Hate Speech biased at eight epochs, where a run did not complete).}
\label{tab:epochs}
\end{table}

\section{Additional results}
\label{sec:appendix-results}

\begin{figure}[htb]
\centering
\includegraphics[width=\textwidth]{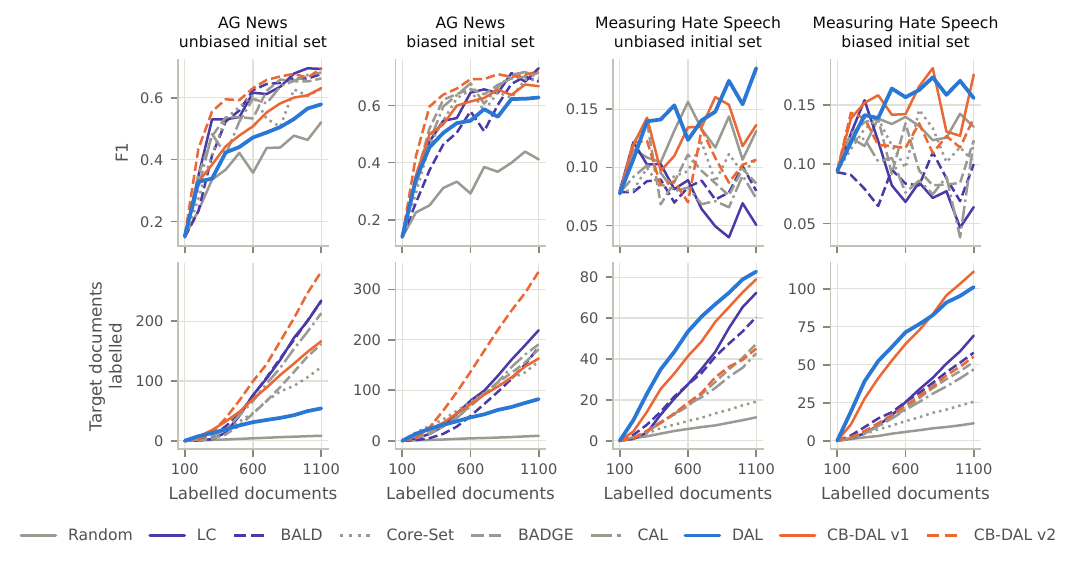}
\caption{Test F1 after each round (top) and target-class documents selected for labelling, cumulatively over rounds (bottom), mean over 25 runs. The bottom row excludes the 50 target-class documents in the initial set.}
\label{fig:curves}
\end{figure}

Table~\ref{tab:main-se} repeats Table~\ref{tab:main} with standard errors. Table~\ref{tab:precision-recall} reports the precision and recall of the target class, and Table~\ref{tab:subgroup-recall} reports recall averaged over subgroups and on the worst subgroup. Figure~\ref{fig:curves} shows test F1 after each round and the number of target-class documents selected for labelling.

\begin{table}[htb]
\centering
\small
\setlength{\tabcolsep}{3pt}
\begin{tabular}{@{}lc@{\hspace{10pt}}ccc@{\hspace{18pt}}c@{\hspace{10pt}}ccc@{}}
\toprule
 & \multicolumn{4}{c}{AG News} & \multicolumn{4}{c}{Measuring Hate Speech} \\
\cmidrule(lr){2-5}\cmidrule(l){6-9}
 & Unbiased & \multicolumn{3}{c}{Biased} & Unbiased & \multicolumn{3}{c}{Biased} \\
\cmidrule(lr){2-2}\cmidrule(lr){3-5}\cmidrule(lr){6-6}\cmidrule(l){7-9}
Method & F1 & F1 & $R_U$ & $R_R$ & F1 & F1 & $R_U$ & $R_R$ \\
\midrule
Random & 0.52{\scriptsize$\pm$0.01} & 0.41{\scriptsize$\pm$0.02} & 0.56{\scriptsize$\pm$0.06} & 0.78{\scriptsize$\pm$0.02} & 0.13{\scriptsize$\pm$0.02} & 0.13{\scriptsize$\pm$0.01} & 0.17{\scriptsize$\pm$0.03} & 0.49{\scriptsize$\pm$0.05} \\
\midrule
LC & 0.69{\scriptsize$\pm$0.01} & 0.73{\scriptsize$\pm$0.01} & 0.90{\scriptsize$\pm$0.02} & 0.68{\scriptsize$\pm$0.02} & 0.05{\scriptsize$\pm$0.02} & 0.06{\scriptsize$\pm$0.02} & 0.02{\scriptsize$\pm$0.01} & 0.10{\scriptsize$\pm$0.03} \\
BALD & 0.68{\scriptsize$\pm$0.01} & 0.68{\scriptsize$\pm$0.03} & 0.87{\scriptsize$\pm$0.04} & 0.65{\scriptsize$\pm$0.04} & 0.08{\scriptsize$\pm$0.02} & 0.10{\scriptsize$\pm$0.02} & 0.09{\scriptsize$\pm$0.03} & 0.23{\scriptsize$\pm$0.05} \\
\midrule
Core-Set & 0.63{\scriptsize$\pm$0.03} & 0.69{\scriptsize$\pm$0.01} & 0.90{\scriptsize$\pm$0.01} & 0.63{\scriptsize$\pm$0.03} & 0.11{\scriptsize$\pm$0.02} & 0.10{\scriptsize$\pm$0.02} & 0.16{\scriptsize$\pm$0.04} & 0.30{\scriptsize$\pm$0.06} \\
BADGE & 0.66{\scriptsize$\pm$0.02} & 0.71{\scriptsize$\pm$0.01} & 0.91{\scriptsize$\pm$0.01} & 0.65{\scriptsize$\pm$0.03} & 0.09{\scriptsize$\pm$0.02} & 0.12{\scriptsize$\pm$0.02} & 0.10{\scriptsize$\pm$0.02} & 0.22{\scriptsize$\pm$0.04} \\
CAL & 0.68{\scriptsize$\pm$0.01} & 0.70{\scriptsize$\pm$0.01} & 0.86{\scriptsize$\pm$0.03} & 0.62{\scriptsize$\pm$0.04} & 0.07{\scriptsize$\pm$0.02} & 0.12{\scriptsize$\pm$0.02} & 0.14{\scriptsize$\pm$0.03} & 0.20{\scriptsize$\pm$0.04} \\
DAL & 0.58{\scriptsize$\pm$0.01} & 0.63{\scriptsize$\pm$0.01} & 0.91{\scriptsize$\pm$0.01} & 0.76{\scriptsize$\pm$0.02} & 0.19{\scriptsize$\pm$0.01} & 0.16{\scriptsize$\pm$0.02} & 0.55{\scriptsize$\pm$0.05} & 0.53{\scriptsize$\pm$0.05} \\
\midrule
CB-DAL v1 & 0.63{\scriptsize$\pm$0.01} & 0.67{\scriptsize$\pm$0.01} & 0.91{\scriptsize$\pm$0.02} & 0.68{\scriptsize$\pm$0.02} & 0.14{\scriptsize$\pm$0.02} & 0.18{\scriptsize$\pm$0.02} & 0.37{\scriptsize$\pm$0.05} & 0.34{\scriptsize$\pm$0.05} \\
CB-DAL v2 & 0.70{\scriptsize$\pm$0.01} & 0.72{\scriptsize$\pm$0.01} & 0.90{\scriptsize$\pm$0.02} & 0.63{\scriptsize$\pm$0.02} & 0.11{\scriptsize$\pm$0.02} & 0.14{\scriptsize$\pm$0.01} & 0.12{\scriptsize$\pm$0.03} & 0.29{\scriptsize$\pm$0.04} \\
\bottomrule
\end{tabular}
\caption{The results in Table~\ref{tab:main}, with standard errors over the 25 runs.}
\label{tab:main-se}
\end{table}

\begin{table}[htb]
\centering
\small
\setlength{\tabcolsep}{2.5pt}
\begin{tabular}{@{}lcc@{\hspace{6pt}}cc@{\hspace{6pt}}cc@{\hspace{6pt}}cc@{}}
\toprule
 & \multicolumn{2}{c}{AG News, unbiased} & \multicolumn{2}{c}{Measuring Hate Speech, unbiased} & \multicolumn{2}{c}{AG News, biased} & \multicolumn{2}{c}{Measuring Hate Speech, biased} \\
\cmidrule(lr){2-3}\cmidrule(lr){4-5}\cmidrule(lr){6-7}\cmidrule(lr){8-9}
Method & P & R & P & R & P & R & P & R \\
\midrule
Random & 0.42{\scriptsize$\pm$0.02} & 0.73{\scriptsize$\pm$0.02} & 0.08{\scriptsize$\pm$0.01} & 0.35{\scriptsize$\pm$0.05} & 0.32{\scriptsize$\pm$0.02} & 0.67{\scriptsize$\pm$0.03} & 0.11{\scriptsize$\pm$0.01} & 0.31{\scriptsize$\pm$0.04} \\
\midrule
LC & 0.70{\scriptsize$\pm$0.01} & 0.70{\scriptsize$\pm$0.01} & 0.10{\scriptsize$\pm$0.04} & 0.05{\scriptsize$\pm$0.02} & 0.69{\scriptsize$\pm$0.02} & 0.79{\scriptsize$\pm$0.02} & 0.12{\scriptsize$\pm$0.03} & 0.06{\scriptsize$\pm$0.02} \\
BALD & 0.69{\scriptsize$\pm$0.02} & 0.69{\scriptsize$\pm$0.02} & 0.06{\scriptsize$\pm$0.02} & 0.14{\scriptsize$\pm$0.04} & 0.63{\scriptsize$\pm$0.03} & 0.76{\scriptsize$\pm$0.04} & 0.11{\scriptsize$\pm$0.02} & 0.15{\scriptsize$\pm$0.04} \\
\midrule
Core-Set & 0.63{\scriptsize$\pm$0.02} & 0.68{\scriptsize$\pm$0.03} & 0.09{\scriptsize$\pm$0.02} & 0.30{\scriptsize$\pm$0.05} & 0.64{\scriptsize$\pm$0.02} & 0.76{\scriptsize$\pm$0.02} & 0.10{\scriptsize$\pm$0.02} & 0.22{\scriptsize$\pm$0.05} \\
BADGE & 0.68{\scriptsize$\pm$0.02} & 0.69{\scriptsize$\pm$0.03} & 0.10{\scriptsize$\pm$0.02} & 0.13{\scriptsize$\pm$0.03} & 0.67{\scriptsize$\pm$0.02} & 0.78{\scriptsize$\pm$0.02} & 0.12{\scriptsize$\pm$0.02} & 0.15{\scriptsize$\pm$0.03} \\
CAL & 0.68{\scriptsize$\pm$0.02} & 0.69{\scriptsize$\pm$0.01} & 0.10{\scriptsize$\pm$0.04} & 0.11{\scriptsize$\pm$0.03} & 0.69{\scriptsize$\pm$0.02} & 0.74{\scriptsize$\pm$0.03} & 0.12{\scriptsize$\pm$0.02} & 0.17{\scriptsize$\pm$0.03} \\
DAL & 0.49{\scriptsize$\pm$0.02} & 0.74{\scriptsize$\pm$0.02} & 0.13{\scriptsize$\pm$0.01} & 0.55{\scriptsize$\pm$0.04} & 0.51{\scriptsize$\pm$0.02} & 0.83{\scriptsize$\pm$0.01} & 0.09{\scriptsize$\pm$0.01} & 0.54{\scriptsize$\pm$0.05} \\
\midrule
CB-DAL v1 & 0.59{\scriptsize$\pm$0.02} & 0.71{\scriptsize$\pm$0.02} & 0.16{\scriptsize$\pm$0.03} & 0.24{\scriptsize$\pm$0.04} & 0.59{\scriptsize$\pm$0.02} & 0.80{\scriptsize$\pm$0.02} & 0.13{\scriptsize$\pm$0.02} & 0.36{\scriptsize$\pm$0.05} \\
CB-DAL v2 & 0.71{\scriptsize$\pm$0.01} & 0.69{\scriptsize$\pm$0.01} & 0.13{\scriptsize$\pm$0.03} & 0.16{\scriptsize$\pm$0.04} & 0.68{\scriptsize$\pm$0.02} & 0.77{\scriptsize$\pm$0.01} & 0.14{\scriptsize$\pm$0.02} & 0.19{\scriptsize$\pm$0.03} \\
\bottomrule
\end{tabular}
\caption{Precision (P) and recall (R) of the target class after the final round, mean $\pm$ standard error over 25 runs.}
\label{tab:precision-recall}
\end{table}

\begin{table}[htb]
\centering
\small
\setlength{\tabcolsep}{2.5pt}
\begin{tabular}{@{}lcc@{\hspace{6pt}}cc@{\hspace{6pt}}cc@{\hspace{6pt}}cc@{}}
\toprule
 & \multicolumn{2}{c}{AG News, unbiased} & \multicolumn{2}{c}{Measuring Hate Speech, unbiased} & \multicolumn{2}{c}{AG News, biased} & \multicolumn{2}{c}{Measuring Hate Speech, biased} \\
\cmidrule(lr){2-3}\cmidrule(lr){4-5}\cmidrule(lr){6-7}\cmidrule(lr){8-9}
Method & Macro & Worst & Macro & Worst & Macro & Worst & Macro & Worst \\
\midrule
Random & 0.73{\scriptsize$\pm$0.02} & 0.73{\scriptsize$\pm$0.02} & 0.30{\scriptsize$\pm$0.05} & 0.03{\scriptsize$\pm$0.01} & 0.67{\scriptsize$\pm$0.03} & 0.54{\scriptsize$\pm$0.05} & 0.20{\scriptsize$\pm$0.03} & 0.00{\scriptsize$\pm$0.00} \\
\midrule
LC & 0.70{\scriptsize$\pm$0.01} & 0.70{\scriptsize$\pm$0.01} & 0.05{\scriptsize$\pm$0.02} & 0.00{\scriptsize$\pm$0.00} & 0.79{\scriptsize$\pm$0.02} & 0.68{\scriptsize$\pm$0.02} & 0.04{\scriptsize$\pm$0.01} & 0.00{\scriptsize$\pm$0.00} \\
BALD & 0.69{\scriptsize$\pm$0.02} & 0.69{\scriptsize$\pm$0.02} & 0.12{\scriptsize$\pm$0.03} & 0.01{\scriptsize$\pm$0.01} & 0.76{\scriptsize$\pm$0.04} & 0.64{\scriptsize$\pm$0.04} & 0.10{\scriptsize$\pm$0.03} & 0.00{\scriptsize$\pm$0.00} \\
\midrule
Core-Set & 0.68{\scriptsize$\pm$0.03} & 0.68{\scriptsize$\pm$0.03} & 0.32{\scriptsize$\pm$0.06} & 0.10{\scriptsize$\pm$0.04} & 0.76{\scriptsize$\pm$0.02} & 0.63{\scriptsize$\pm$0.03} & 0.16{\scriptsize$\pm$0.04} & 0.00{\scriptsize$\pm$0.00} \\
BADGE & 0.69{\scriptsize$\pm$0.03} & 0.69{\scriptsize$\pm$0.03} & 0.10{\scriptsize$\pm$0.03} & 0.01{\scriptsize$\pm$0.01} & 0.78{\scriptsize$\pm$0.02} & 0.65{\scriptsize$\pm$0.03} & 0.11{\scriptsize$\pm$0.02} & 0.00{\scriptsize$\pm$0.00} \\
CAL & 0.69{\scriptsize$\pm$0.01} & 0.69{\scriptsize$\pm$0.01} & 0.12{\scriptsize$\pm$0.03} & 0.01{\scriptsize$\pm$0.01} & 0.74{\scriptsize$\pm$0.03} & 0.61{\scriptsize$\pm$0.04} & 0.15{\scriptsize$\pm$0.03} & 0.01{\scriptsize$\pm$0.01} \\
DAL & 0.74{\scriptsize$\pm$0.02} & 0.74{\scriptsize$\pm$0.02} & 0.60{\scriptsize$\pm$0.04} & 0.22{\scriptsize$\pm$0.04} & 0.83{\scriptsize$\pm$0.01} & 0.75{\scriptsize$\pm$0.02} & 0.55{\scriptsize$\pm$0.06} & 0.17{\scriptsize$\pm$0.04} \\
\midrule
CB-DAL v1 & 0.71{\scriptsize$\pm$0.02} & 0.71{\scriptsize$\pm$0.02} & 0.24{\scriptsize$\pm$0.05} & 0.05{\scriptsize$\pm$0.02} & 0.79{\scriptsize$\pm$0.02} & 0.68{\scriptsize$\pm$0.02} & 0.33{\scriptsize$\pm$0.05} & 0.05{\scriptsize$\pm$0.02} \\
CB-DAL v2 & 0.69{\scriptsize$\pm$0.01} & 0.69{\scriptsize$\pm$0.01} & 0.17{\scriptsize$\pm$0.04} & 0.02{\scriptsize$\pm$0.01} & 0.77{\scriptsize$\pm$0.01} & 0.63{\scriptsize$\pm$0.02} & 0.14{\scriptsize$\pm$0.03} & 0.01{\scriptsize$\pm$0.01} \\
\bottomrule
\end{tabular}
\caption{Recall averaged over the subgroups of the target class (Macro) and on the subgroup with the lowest recall (Worst) after the final round, mean $\pm$ standard error over 25 runs. With an unbiased initial set, the AG News target class has a single subgroup, so both equal recall. In Measuring Hate Speech the smallest subgroup has one test document, so worst-group recall is coarse.}
\label{tab:subgroup-recall}
\end{table}

\end{document}